\documentclass[10pt,twocolumn,letterpaper]{article}

\usepackage{cvpr}              
\definecolor{cvprblue}{rgb}{0.21,0.49,0.74}
\usepackage[pagebackref,breaklinks,colorlinks,allcolors=cvprblue]{hyperref}
\usepackage[utf8]{inputenc} 
\usepackage[T1]{fontenc}    
\usepackage{hyperref}       
\usepackage{url}            
\usepackage{booktabs}       
\usepackage{amsfonts}       
\usepackage{nicefrac}       
\usepackage{microtype}      
\usepackage{xcolor}         

\usepackage[utf8]{inputenc} 
\usepackage[T1]{fontenc}    
\usepackage{hyperref}       
\hypersetup{
    colorlinks=true,
    citecolor=green
}
\usepackage{url}            
\usepackage{booktabs}       
\usepackage{amsfonts}       
\usepackage{xspace}
\usepackage{nicefrac}       
\usepackage{microtype}      
\usepackage{xcolor}         
\usepackage{graphicx}
\usepackage{url}            
\usepackage{booktabs}       
\usepackage{amsfonts}       
\usepackage{nicefrac}       
\usepackage{microtype}      
\usepackage{xcolor}         
\usepackage{graphicx}
\usepackage{wasysym}
\usepackage{multirow}
\usepackage{tcolorbox}
\usepackage[table]{xcolor}
\usepackage{subcaption}
\usepackage{amsfonts}
\usepackage{pifont}
\usepackage{fontawesome}
\usepackage{float}
 \usepackage{enumitem}
\usepackage[dvipsnames]{xcolor}
\usepackage[normalem]{ulem}
\usepackage{xspace}
\usepackage{makecell}
\usepackage{wrapfig}
\usepackage{xcolor}
\usepackage{colortbl}
\usepackage{multirow}
\usepackage[table]{xcolor}
\usepackage{tcolorbox}

\definecolor{softblue}{RGB}{230,245,255}
\definecolor{softgreen}{RGB}{220,245,220}
\definecolor{softyellow}{HTML}{FFF2CC}
\definecolor{softpurple}{RGB}{245,235,255}
\definecolor{softorange}{RGB}{255,235,210}
\definecolor{softpink}{RGB}{255,230,240}
\definecolor{lightgray}{RGB}{240,240,240}

\newcommand{\icoyes}{\textcolor{ForestGreen}{\ding{51}}\xspace} 
\newcommand{\icono}{\textcolor{Red}{\ding{55}}\xspace}          

\newcommand{\name}{Artifact-Bench}

\def\paperID{*****} 
\def\confName{CVPR}
\def\confYear{2026}

\title{Artifact-Bench: Evaluating MLLMs on Detecting and Assessing the\\ Artifacts of AI-Generated Videos}

\author{%
    Yuqi Tang$^{1,3}$\thanks{Equal Contribution} \,\,
    Yang Shi$^{2,3}$\footnotemark[1]\,\,\thanks{Project Lead} \,\,
    Zhuoran Zhang$^{2}$\footnotemark[1]\,\,\,
    Qixun Wang$^{2}$\footnotemark[1]\,\,\,
    Xuehai Bai$^{4}$\,
    Yue Ding$^{5}$\,
    \\
    Ruizhe Chen$^{6}$\,
    Bohan Zeng$^{2}$\,
    Xinlong Chen$^{5}$\,
    Xuanyu Zhu$^{2}$\,
    Bozhou Li$^{2}$\,
    Yuran Wang$^{2}$\,
    Yifan Dai$^{7}$\,
    \\
    Chengzhuo Tong$^{2}$\,
    Xinyu Liu$^{8}$\,
    Yiyan Ji$^{9}$\,
    Yujie Wei$^{10}$\,
    Yuhao Dong$^{11}$\,
    Shilin Yan$^{10}$\,
    \\
    Fengxiang Wang$^{12}$\,
    Yi-Fan Zhang$^{5}$\thanks{Corresponding Author}\,\,\,
    Haotian Wang$^{13}$\footnotemark[3]\,\,\,
    Yuanxing Zhang$^{3}$\footnotemark[3]\,\,\,
    Pengfei Wan$^{3}$
    \\ 
    $^1$HKUST(GZ)\quad
    $^2$PKU\quad
    $^3$Kling Team\quad
    $^4$HDU\quad
    $^5$CASIA\quad
    $^6$ZJU\quad
    $^{7}$SJTU\quad
    \\
    $^{8}$HKUST\quad
    $^{9}$NJU\quad
    $^{10}$FDU\quad
    $^{11}$NTU\quad
    $^{12}$Shanghai AI Lab\quad
    $^{13}$THU\quad
    \\
    {\centering}
    \url{https://github.com/FrankYang-17/Artifact-Bench}
}

\begin{document}
\maketitle

\begin{abstract}
Recent video generative models have greatly improved the realism of AI-generated videos, yet their outputs still exhibit artifacts such as temporal inconsistencies, structural distortions, and semantic incoherence. 
While Multimodal Large Language Models (MLLMs) show strong visual understanding capabilities, their ability to perceive and reason about such artifacts remains unclear. 
Existing benchmarks often lack systematic evaluation of artifact-aware perception and fine-grained diagnostic reasoning, especially across diverse AI-generated video domains beyond photorealistic content. 
To address this gap, we introduce \textbf{\name}, a comprehensive benchmark for evaluating MLLMs on AI-generated video artifact detection and analysis. 
We first establish a three-level hierarchical taxonomy of realism artifacts, covering photorealistic, animated, and CG-style videos. 
Based on this taxonomy, \name\ defines three complementary tasks: real vs.\ AI-generated video classification, pairwise realism comparison, and fine-grained artifact identification. 
Experiments on $19$ leading MLLMs reveal substantial limitations in artifact perception and reasoning, with many models approaching random or even below-random performance in challenging settings. 
We further observe significant misalignment between MLLM judgments and human perceptual preferences, highlighting their limited reliability as general evaluators for AI-generated video realism.
\end{abstract}

\section{Introduction} 

Recent advances in video generative models~\cite{google2025veo3,Kling, davinci-magihuman-2026,wan2025,LTX-2,hunyuan-1.5} have significantly improved the quality of AI-generated videos, enabling the synthesis of visually compelling content with increasingly realistic appearance and motion. 
Despite this progress, most generated videos still exhibit noticeable imperfections, such as temporal inconsistencies, structural distortions, unnatural motion, and semantic incoherence. 
These artifacts, although sometimes subtle, fundamentally limit perceptual realism and hinder reliable deployment in real-world applications~\cite{Skyra,VideoVeritas}.


Distinguishing AI-generated videos from real-world ones has therefore become increasingly important for media authenticity, content moderation, and generative model evaluation.
Among various cues, generative artifacts provide particularly informative signals, as they often reflect intrinsic limitations of current generation pipelines rather than high-level semantics.
Compared to purely semantic or style-based cues, artifact-based detection offers a more principled pathway for identifying AI-generated content~\cite{Skyra,VideoVeritas}, especially as generative models continue to improve in visual fidelity.
Beyond binary classification, an underexplored question is whether models can identify and diagnose these artifacts, enabling more interpretable judgments and providing insights for improving generative models.
In this sense, artifact analysis serves as a critical bridge between evaluation and generation, facilitating the refinement of video generation systems toward higher realism.

\begin{table*}[t]
\centering
\small 
\setlength{\tabcolsep}{3.5pt}
\caption{
\textbf{Comprehensive Comparison with Other Benchmarks.}
\name features a multi-granularity progressive three-task system with difficulty levels, systematically evaluating model capabilities in AIGC video detection, realism comparison, and fine-grained artifact diagnosis across comprehensive scenarios (including non-photorealistic types such as CG and animation) and a well-established artifact taxonomy of 30 evaluation aspects.}
\label{tab:comparison}
\vspace{-0.5em}
\resizebox{\textwidth}{!}{ 
\begin{tabular}{lcccccccccc}
\toprule
\multirow{2}{*}{\textbf{Benchmark}} & \multirow{2}{*}{\textbf{\#Tasks}} & \multirow{2}{*}{\textbf{\#Videos}} & \multirow{2}{*}{\textbf{\#Samples}} & \multirow{2}{*}{\textbf{Scenarios}} & \multirow{2}{*}{\makecell[c]{\textbf{Diff.} \\ \textbf{Levels}}} & \multicolumn{3}{c}{\textbf{Multi-granularity}} & \multicolumn{2}{c}{\textbf{Annotation}} \\
\cmidrule(lr){7-9} \cmidrule(lr){10-11}
& & & & & & \textbf{Det.} & \textbf{Comp.} & \textbf{Diag.} & \textbf{Annotator} & \textbf{Eval. Aspects} \\
\midrule
ViF-Bench~\cite{Skyra} & 1 & 2,995 & 2,995 & Real & \icono & \icoyes & \icono & \icoyes & Human+MLLM & 23 \\
GenBuster-Bench~\cite{Busterx} & 2 & 3,150 & 3,150 & Real & \icoyes & \icoyes & \icono & \icoyes & MLLM & 3 \\
VF-Eval~\cite{vfeval} & 4 & 9,740 & 9,740 & Real \& Stylized & \icono & \icono & \icono & \icoyes & Human+MLLM & 11 \\
UVE-Bench~\cite{uve} & 2 & 1,230 & 4,042 & Real \& Stylized & \icono & \icono & \icoyes & \icoyes & Human & 15 \\
AEGIS~\cite{AEGIS} & 1 & 3,166 & 3,166 & Real & \icoyes & \icoyes & \icono & \icoyes & MLLM & 3 \\
\midrule
\rowcolor{gray!10} \textbf{Artifact-Bench} & \textbf{3} & \textbf{1,350} & \textbf{1,100} & \textbf{Real \& Stylized} & \icoyes & \icoyes & \icoyes & \icoyes & \textbf{Human} & \textbf{30} \\
\bottomrule
\multicolumn{11}{l}{\footnotesize *Det., Comp., and Diag. denote AIGC Video Detection, Realism Comparison, and Artifact Diagnosis tasks, respectively.}
\end{tabular}
} 
\vspace{-2em}
\end{table*}

In parallel, Multimodal Large Language Models (MLLMs)~\cite{qwen3-vl,mavors,gemini-3.1-pro,gpt-4.1,wang2026geollava,wang2026text,wang2026geoeyes} have emerged as powerful general-purpose models for visual reasoning. 
Their ability to process complex visual inputs and generate structured language outputs makes them promising candidates for scalable video evaluation. 
However, it remains unclear whether current MLLMs can genuinely perceive and reason about AIGC-specific artifacts. 
As shown in Table~\ref{tab:comparison}, existing benchmarks have explored authenticity detection, preference evaluation, and artifact grounding, but often in isolated settings or limited photorealistic scenarios. 
Moreover, most video benchmarks emphasize semantic understanding and general reasoning rather than perceptual realism and generative artifacts, making it difficult to determine whether MLLMs rely on genuine artifact-aware perception or superficial semantic priors and dataset biases.

To address this gap, we first conduct a systematic analysis of common artifacts in AI-generated videos, covering their characteristics, causes, and perceptual manifestations. 
Based on this analysis, we establish a three-level artifact taxonomy that organizes AIGC video artifacts from coarse visual abnormalities to fine-grained structural and temporal inconsistencies, providing a principled foundation for artifact-oriented evaluation. 
Building on this taxonomy, we introduce \textbf{\name}, a benchmark for evaluating MLLMs on AI-generated video artifact detection and analysis. 
\name\ consists of three complementary tasks: real vs.\ AI-generated video classification, pairwise realism comparison, and fine-grained artifact identification, which progressively probe model capabilities from coarse-grained recognition to diagnostic reasoning. 
To support reliable evaluation, we develop a hybrid data construction pipeline combining real-world video collection, controlled generation, and targeted artifact synthesis, together with a difficulty stratification scheme that captures varying levels of realism and artifact subtlety.


Extensive experiments on \name\ reveal fundamental limitations of current MLLMs in perceiving and understanding artifacts in AI-generated videos. 
Despite strong general vision-language capabilities, many models show near-random or even below-random performance on certain tasks, exposing severe weaknesses in artifact-level perception and reasoning. 
Moreover, model judgments often misalign with human perceptual preferences and do not consistently follow the human-defined difficulty hierarchy, suggesting reliance on superficial statistical cues or semantic priors rather than genuine artifact perception. 
These findings show that artifact-aware perception remains far from solved and call for future MLLMs with stronger human-aligned realism understanding and fine-grained perceptual reasoning.

We summarize our main contributions as follows:
\begin{enumerate}[leftmargin=*, itemsep=1pt, parsep=0pt, topsep=2pt]
    \item We conduct a systematic study of artifacts in AI-generated videos and establish a three-level hierarchical taxonomy that organizes AIGC-specific artifacts from coarse visual abnormalities to fine-grained temporal and structural inconsistencies, providing a principled foundation for artifact-aware evaluation and analysis.
    \item We introduce \textbf{\name}, a comprehensive benchmark for evaluating the ability of MLLMs to detect and analyze artifacts in AI-generated videos. Based on our artifact taxonomy, we design a multi-level evaluation framework consisting of three complementary tasks: real vs. AI-generated video classification, pairwise realism comparison, and fine-grained artifact identification. We further develop a hybrid data construction pipeline with carefully designed difficulty stratification to support reliable and in-depth evaluation.
    \item We conduct extensive experiments across a diverse set of state-of-the-art MLLMs and reveal fundamental limitations of current models in artifact-level perception and reasoning. Our findings show that many MLLMs exhibit near-random or even below-random performance on challenging tasks and demonstrate significant misalignment with human perceptual preferences, highlighting the urgent need for future MLLMs with stronger human-aligned realism understanding capabilities.
\end{enumerate}

\section{Related Work}


\subsection{Multimodal Large Language Model}

Multimodal Large Language Models (MLLMs)~\cite{gemini-3.1-pro,qwen3-vl,internvl3.5,gpt-4o,mavors, zhang2025mm,MiniCPM-V-4.5,glm4.6-v} have recently demonstrated remarkable proficiency in visual understanding and multimodal reasoning. 
Specifically, their capacity to process and interpret temporal information has enabled a diverse array of video-based applications, such as visual question answering~\cite{Versavid-r1, debiasingmllm}, video captioning \cite{mavors,Avocado}, and video-based optical character recognition (OCR) \cite{MiniCPM-V-4.5,MME-VideoOCR}. 
Beyond basic perception, MLLMs excel in complex visual reasoning \cite{Monet, chen2025opengpt,zhang2025modalities,bai2026edit}, making them increasingly viable for sophisticated real-world scenarios~\cite{Embodiedeval, VTC-Bench}. 
Leveraging these robust capabilities, recent research has begun to explore MLLMs for automated AI-generated video detection and realism assessment, as exemplified by works like BusterX++~\cite{Busterx++} and Skyra \cite{Skyra}.



\subsection{Benchmarks for AI-Generated Video Detection and Assessment}

As video generative models continue to advance, recent studies have explored MLLMs as general-purpose tools for detecting and assessing artifacts in AI-generated videos. 
Some benchmarks focus on quality assessment and diagnostic feedback. 
UVE-Bench~\cite{uve} introduces pairwise comparison scoring across fine-grained dimensions with human preference annotations, while VF-Eval~\cite{vfeval} formulates evaluation as a diagnostic Question-Answering (QA) task. 
However, preference-based scoring provides limited insight into model reasoning, and QA-style evaluation may allow models to exploit dataset biases. 
Other benchmarks focus on authenticity detection and artifact localization. 
AEGIS~\cite{AEGIS} provides multi-modality feature annotations to evaluate model reasoning chains, GenBuster-Bench~\cite{Busterx} adopts an MLLM-as-a-Judge protocol to assess authenticity prediction rationales, and ViF-Bench~\cite{Skyra} requires spatial-temporal grounding with timestamps and bounding boxes based on a hierarchical artifact taxonomy.
Despite these advances, existing benchmarks remain limited in two aspects. 
First, they typically evaluate models under a single paradigm, such as authenticity classification, preference scoring, or artifact grounding, lacking a unified multi-granularity evaluation framework. 
Second, their evaluation scenarios are often narrow, primarily focusing on photorealistic AI-generated videos. 
In contrast, {\name} introduces three progressively challenging tasks: real vs.\ AI-generated video classification, pairwise realism comparison, and fine-grained artifact identification. 
These tasks systematically evaluate MLLMs from coarse authenticity perception to fine-grained artifact reasoning. 
Moreover, {\name} covers diverse video domains, including photorealistic, anime, and CG-style videos, offering broader applicability and stronger practical relevance.


\section{\name}


\subsection{Taxonomy of Realism Artifacts in AI-Generated Videos}
\begin{figure*}
    \centering
    \includegraphics[width=1\linewidth]{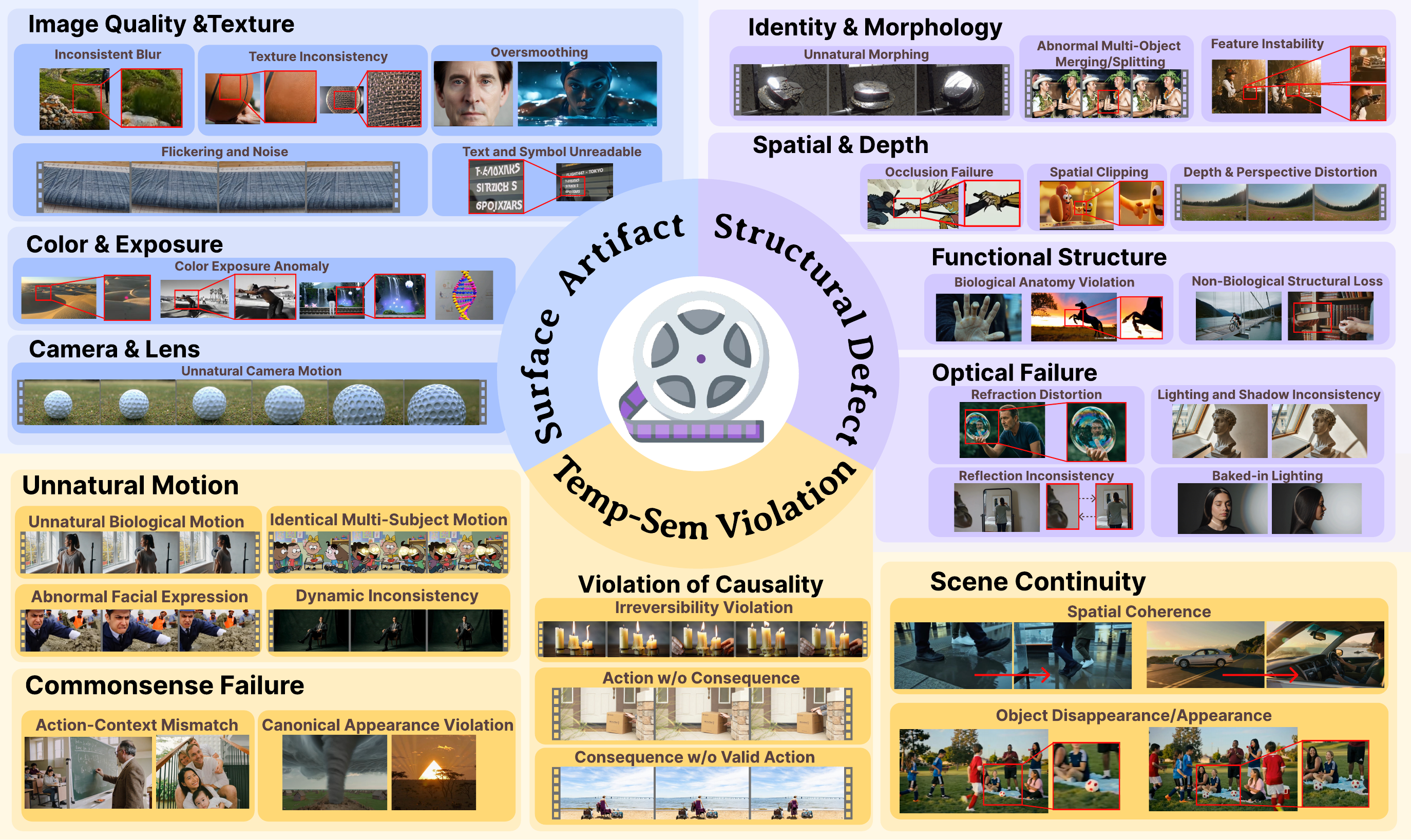}
    \caption{
    \textbf{The Hierarchical Taxonomy of AI-Generated Video Artifacts.} 
    We organize AI-generated video realism artifacts into three hierarchical tiers: top-level artifact domains, mid-level failure families, and 30 fine-grained artifact types. 
    The taxonomy spans Surface Artifacts, Structural Defects, and Temporal-Semantic (Temp-Sem) Violations, covering failures in visual appearance, object and scene structure, temporal continuity, causality, and semantic coherence. 
    Annotated visual examples are provided for representative fine-grained artifact types.
    }
    \label{fig:taxonomy}
\end{figure*}

To support fine-grained evaluation of MLLMs on AI-generated video realism, we first establish a hierarchical taxonomy of realism artifacts. 
Unlike general video quality degradation or artifacts introduced by traditional rendering pipelines, artifacts in AI-generated videos often arise from the limitations of generative models in maintaining visual fidelity, object structure, temporal continuity, and semantic consistency. 
These artifacts provide important evidence for distinguishing AI-generated videos from real-world ones and, more importantly, for explaining why a generated video appears unrealistic. We construct the taxonomy through an iterative human analysis process. 
Specifically, we examine a diverse collection of publicly accessible AIGC videos, including photorealistic videos, stylized videos, and computer-generated visuals that aim to simulate realistic appearance or motion. 
By repeatedly inspecting these videos, identifying recurring failure patterns, and merging semantically overlapping cases, we iteratively refine the category boundaries and ultimately establish a hierarchical taxonomy, as shown in Figure~\ref{fig:taxonomy}. 

The taxonomy is designed to cover the major types of artifacts observed in AI-generated videos as comprehensively as possible, while keeping each category interpretable and actionable for human annotation and model evaluation. 
It is organized into three hierarchical tiers, progressing from broad artifact domains to fine-grained diagnostic labels.

At the highest tier, we divide realism artifacts into three top-level artifact domains according to the perceptual and reasoning depth required for detection. 
\textit{Surface Artifacts} refer to low-level visual defects that can be identified primarily from local appearance cues. 
\textit{Structural Defects} capture failures that require understanding the organization of objects and scenes. 
\textit{Temporal-Semantic Violations} represent higher-level failures that require integrating information across frames and applying commonsense or causal reasoning.

The middle tier further decomposes each top-level domain into failure families that describe the source of the underlying defect. 
For instance, within Surface Artifacts, \textit{Color \& Exposure}, \textit{Camera \& Lens}, and \textit{Image Quality \& Texture} represent failures of distinct visual formation or rendering processes. 
Similarly, Structural Defects involve failure families related to identity, morphology, spatial depth, functional structure, and optical consistency, while Temporal-Semantic Violations cover failures in motion, causality, commonsense, and scene continuity. 
This structure allows defects with different physical, geometric, or semantic origins to be diagnosed independently.

The finest tier provides the most fine-grained artifact descriptions and serves as the operational label space for artifact-oriented evaluation. 
It contains 30 fine-grained artifact types, each corresponding to a concrete and visually observable failure mode, such as \textit{Texture Inconsistency}, \textit{Irreversibility Violation}, or \textit{Cross-Shot Coherence}. 

The taxonomy is diagnostic rather than strictly mutually exclusive. 
A single video may contain multiple co-occurring artifacts, and one visible failure may involve multiple levels of analysis, such as structural deformation and temporal inconsistency. 
Therefore, Artifact-Bench supports multi-label artifact annotations, enabling a more faithful evaluation of whether MLLMs can identify the diverse causes of unrealism in AI-generated videos.

\subsection{Benchmark Design}
\begin{figure*}
    \centering
    \includegraphics[width=1\linewidth]{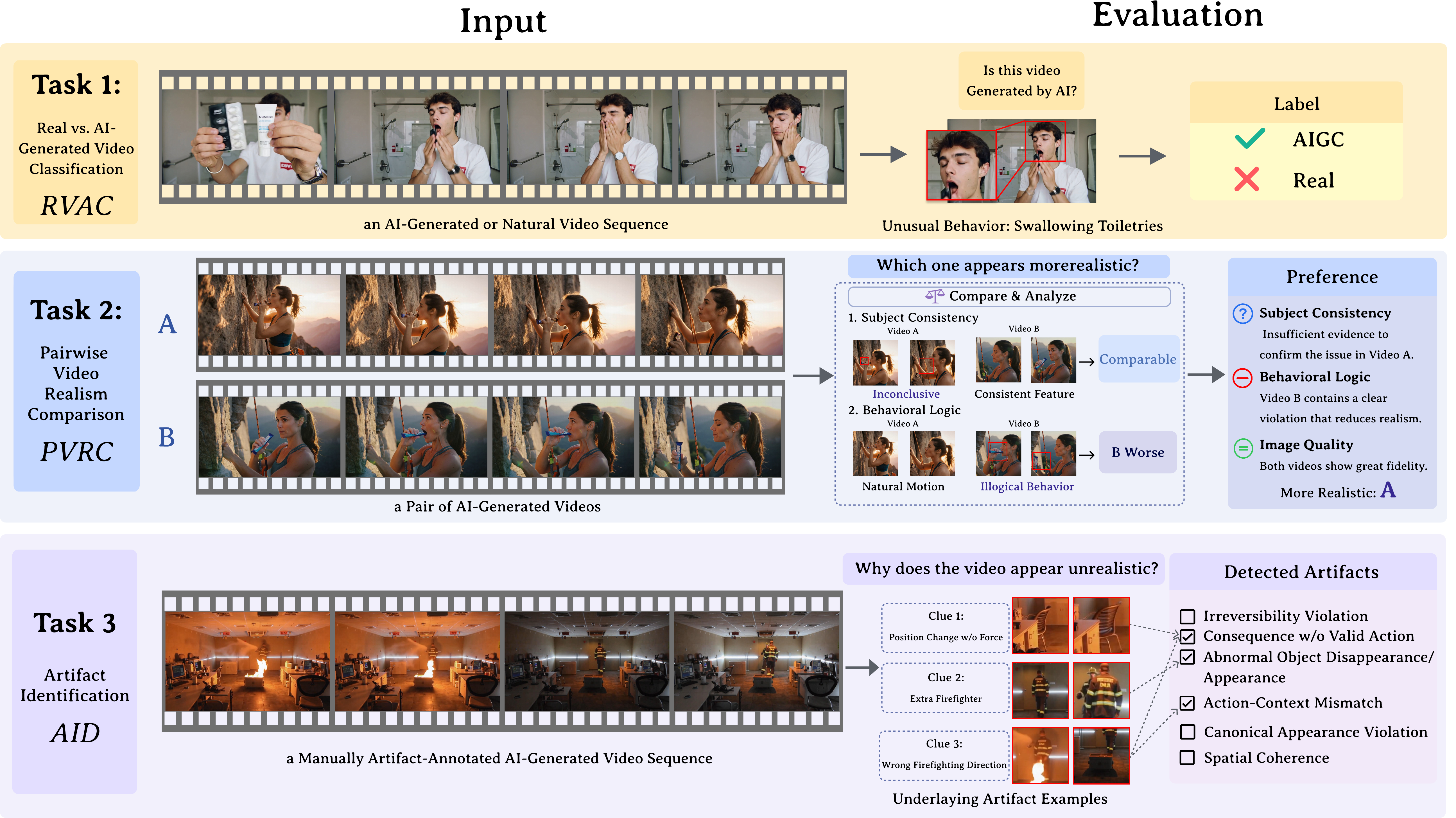}
    \caption{
    \textbf{Illustration of the three proposed tasks and their evaluation workflows.} 
    From top to bottom, the figure demonstrates the input formats and expected reasoning pipelines for RVAC, PVRC, and AID. These tasks form a comprehensive hierarchy, evaluating model capabilities from coarse-grained recognition to detailed artifact identification.
    }
    \label{fig:main}
\end{figure*}

To comprehensively evaluate the capability of MLLMs in recognizing and reasoning about AI-generated videos, we design $3$ complementary tasks in \name (as illustrated in Figure~\ref{fig:main}). 
These tasks progressively evaluate different aspects of authenticity understanding, including (1) distinguishing AI-generated videos from real ones, (2) comparing the realism of different synthetic videos, and (3) identifying specific artifacts that reduce video realism. 
Together, these tasks provide a multi-level assessment of model capabilities ranging from coarse-grained recognition to fine-grained reasoning.

\noindent\textbf{Task 1: Real vs. AI-Generated Video Classification (RVAC).}
This task evaluates the ability of MLLMs to recognize AI-generated videos. 
Given a single video as input, the model must determine whether the video is real or AI-generated and output a binary answer (``\texttt{Yes}'' or ``\texttt{No}'') indicating whether the video is synthetic. 
Each real video in the task is paired with an AI-generated counterpart that shares similar semantic content, ensuring that the task focuses on identifying realism-related artifacts rather than semantic differences. 
This task primarily measures whether MLLMs can detect visual inconsistencies commonly observed in generated videos, such as abnormal motion patterns, implausible physical interactions, or temporal incoherence.

\noindent\textbf{Task 2: Pairwise Video Realism Comparison (PVRC).}
Beyond recognizing AI-generated videos, the second task evaluates whether MLLMs can assess the relative realism of synthetic videos. 
Specifically, the model is given two AI-generated videos (\texttt{<video A>} and \texttt{<video B>}) and must select the one that appears more realistic by responding with either ``\texttt{video A}'' or ``\texttt{video B}''. 
The two videos in each pair share similar semantic content, ensuring that the comparison focuses on differences in visual realism rather than scene semantics. 
Compared with binary classification, this pairwise formulation provides a more fine-grained evaluation of a model's ability to judge the relative realism of AI-generated videos.

\noindent\textbf{Task 3: Artifact Identification (AID).}
This task further evaluates the fine-grained reasoning ability of MLLMs in accurately identifying artifacts in AI-generated videos, requiring models to explain why a video appears unrealistic. 
Given an AI-generated video, the model is asked to determine the primary cause of its unrealism. 
Each example is formulated as a multi-answer multiple-choice question with $6$ candidate options, all of which are instantiated from the 30 fine-grained artifact types in our taxonomy. 
The correct options correspond to the fine-grained artifact labels that are clearly observable in the video. 
The incorrect options are selected from semantically related or visually confusable artifact types, typically within the same or adjacent failure families. 
This design prevents models from solving the task through coarse category elimination and instead requires them to discriminate among fine-grained causes of unrealism. 
The model is required to select all valid fine-grained artifact labels from the $6$ candidates.
By requiring explicit identification of the underlying artifact, this task provides a deeper evaluation of whether MLLMs can analyze and reason about the causes of visual unrealism rather than merely recognizing synthetic content.


\subsection{Benchmark Construction}

\begin{figure*}
    \centering
    \includegraphics[width=\linewidth]{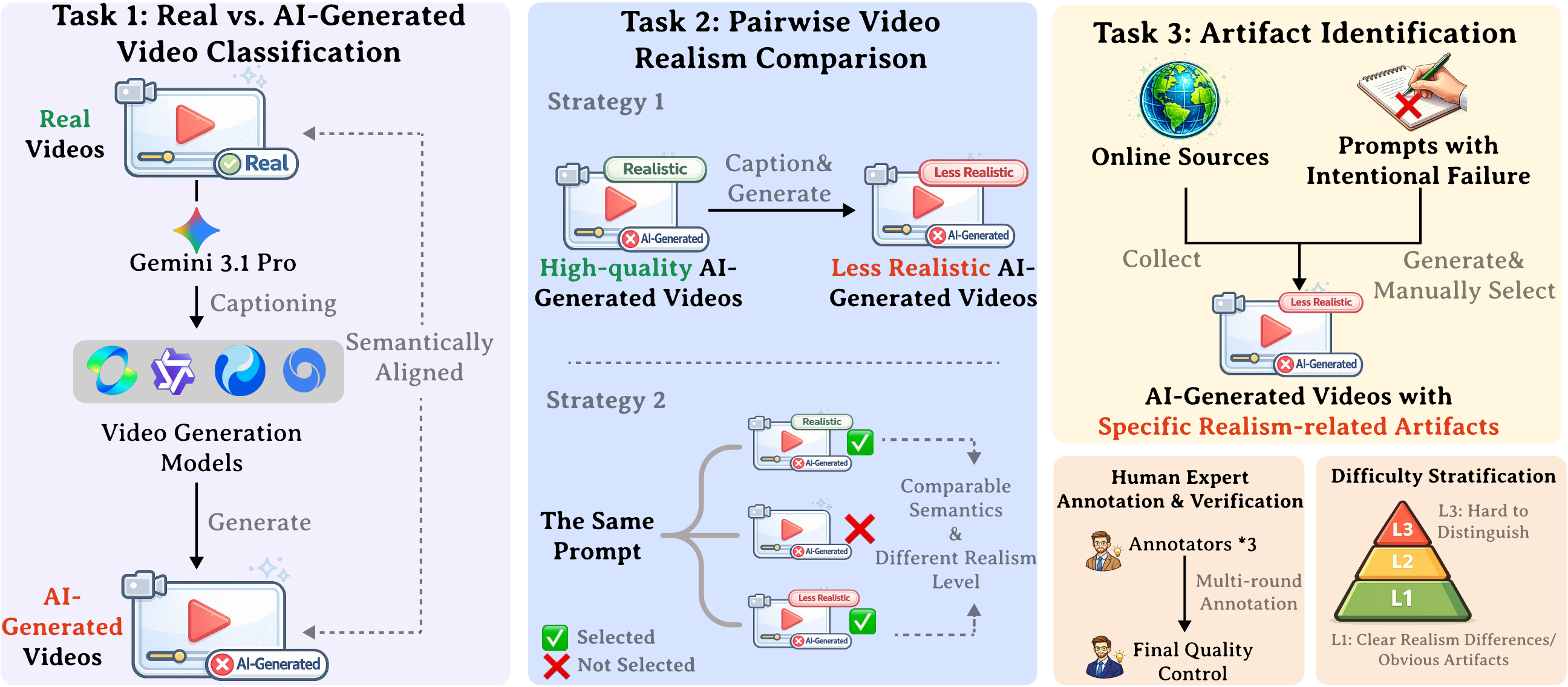}
    \caption{
    \textbf{Overview of the \name~construction pipeline.} We build a hybrid dataset combining real-world and AI-generated videos for three tasks: RVAC, PVRC, and AID. Real videos are captioned to generate semantically aligned AIGC counterparts, while AIGC pairs with varying realism are constructed via re-generation and prompt-consistent sampling. For artifact coverage, we combine natural collection with targeted generation. All samples are manually annotated, verified, and stratified into three difficulty levels (L1–L3) based on realism and artifact severity.
    }
    \label{fig:benchmark_construction}
\end{figure*}

\begin{figure*}
    \centering
    \includegraphics[width=1\linewidth]{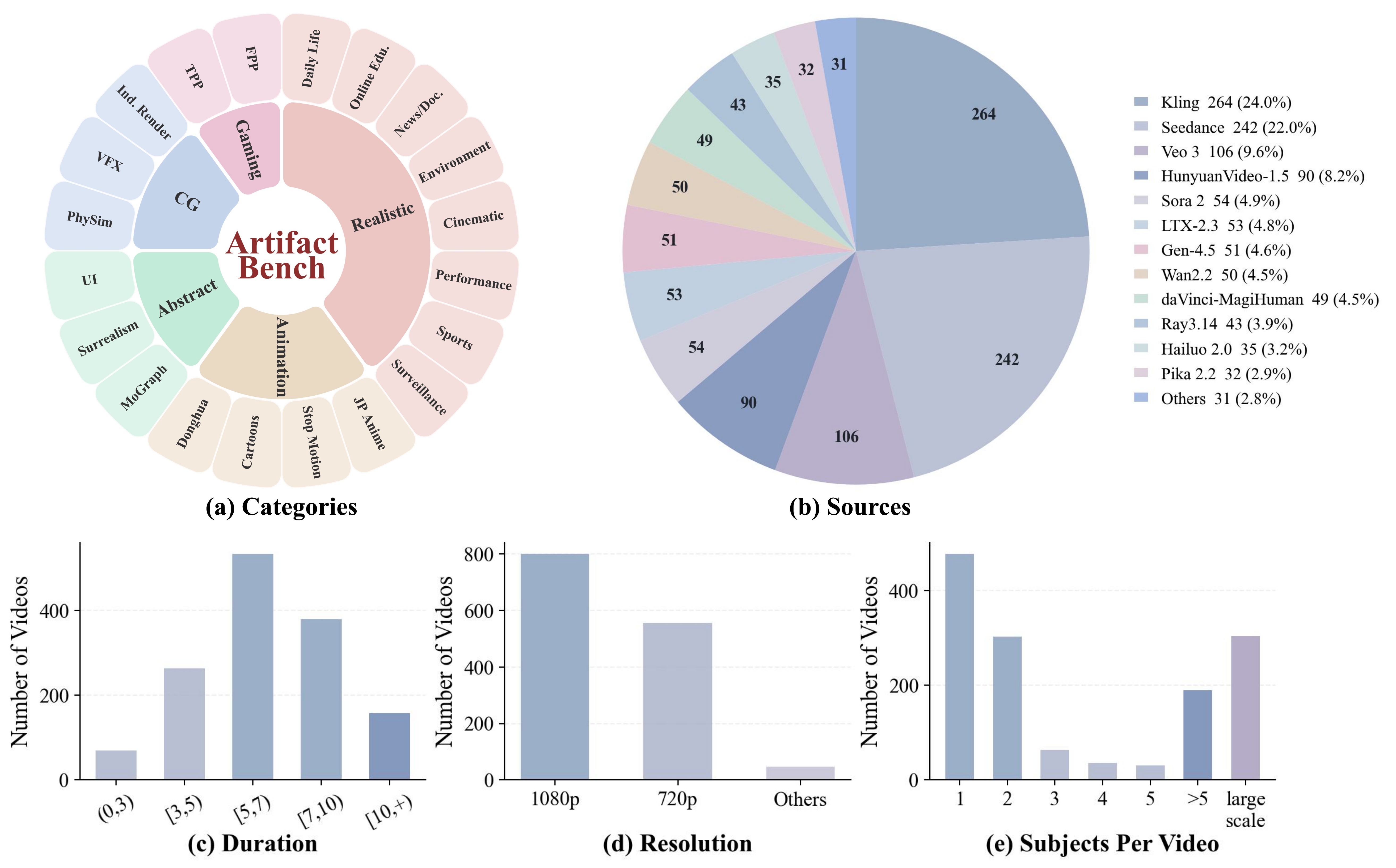}
    \caption{
    \textbf{Statistics of \name.} \textbf{(a)} Hierarchical breakdown of the major video categories and diverse sub-scenarios. \textbf{(b)} Distribution of video sources, featuring a variety of recent state-of-the-art generative models. \textbf{(c)--(e)} Distributions of video duration, spatial resolution, and the number of primary subjects, respectively, demonstrating the structural diversity of our dataset.
    }
    \label{fig:stats}
\end{figure*}

\textbf{Data Collection.}
We construct the benchmark by combining publicly available online videos with model-generated synthetic videos, which enables us to balance semantic controllability, realism diversity, and artifact coverage across different tasks. 
Since the three tasks in \name~target different capabilities, we adopt task-specific data construction pipelines, as shown in Figure~\ref{fig:benchmark_construction}. We use Gemini 3.1 Pro~\cite{gemini-3.1-pro} to generate detailed captions for videos, and employ multiple video generative models to promote diversity in the generated AIGC videos, including Kling-2.5~\cite{Kling}, Kling-2.1~\cite{Kling}, Veo 3~\cite{google2025veo3}, HunyuanVideo-1.5~\cite{hunyuan-1.5}, daVinci-MagiHuman~\cite{davinci-magihuman-2026}, LTX-2.3~\cite{LTX-2}, and Wan2.2~\cite{wan2025}. 

For \textit{Task 1: Real vs. AI-Generated Video Classification (RVAC)}, we first collect and carefully curate real-world videos from publicly available online sources. 
We then caption these videos and use the captions as prompts to generate semantically aligned AI-generated counterparts with video generative models. 
This one-to-one construction ensures semantic alignment, thereby directing the task toward realism-related cues rather than semantic differences.

For \textit{Task 2: Pairwise Video Realism Comparison (PVRC)}, we construct semantically aligned AI-generated video pairs with varying realism levels using two complementary strategies. 
First, we collect high-quality AI-generated videos from publicly available sources, caption them, and use the captions to generate less realistic counterparts. 
Second, we directly generate multiple videos from the same prompt and select pairs with comparable semantics but varying levels of realism and artifact severity. 
Together, these strategies ensure both semantic alignment and sufficient contrast in realism and artifact severity within each pair.

For \textit{Task 3: Artifact Identification (AID)}, we aim to cover a diverse set of realism-related artifacts in AI-generated videos. 
We first collect AIGC videos from online sources that clearly exhibit specific artifact types. 
However, we observe that certain artifacts are rarely present in naturally collected AIGC videos. 
To address this, we design prompts to intentionally expose such failure modes, generate candidate videos, and manually select qualified samples. 
This combination of natural collection and targeted generation improves the coverage and diversity of artifacts in the benchmark.

\noindent\textbf{Annotation and Verification.}
Given that many AI-generated videos are visually close to real-world videos, we adopt a fully manual annotation protocol to ensure reliability. 
Each AI-generated video is independently examined by $3$ experienced annotators, who analyze realism-related artifacts and provide detailed annotations. 
A sample is accepted only if all $3$ annotators reach consistent conclusions; otherwise, it undergoes a second round of review by $2$ additional annotators. 
Finally, all accepted samples are further verified by $2$ expert annotators with extensive industry experience, providing an additional layer of quality control to ensure reliability.

\noindent\textbf{Difficulty Stratification.}
To systematically evaluate model sensitivity to varying levels of realism and artifact severity, we introduce a difficulty stratification scheme over all task samples. Specifically, based on the degree of visual realism, samples are grouped into $3$ levels (L1–L3) with increasing difficulty.
For Task 1 and Task 3, L1 corresponds to low-realism videos with obvious artifacts, making them easy to identify, while L3 consists of highly realistic videos that are difficult to distinguish. 
For Task 2, L1 denotes pairs with clear differences in realism and artifact severity, whereas L3 includes pairs with highly similar realism and subtle artifact patterns, requiring fine-grained perception to differentiate.
To ensure annotation reliability despite the inherent subjectivity of difficulty assessment, each sample is independently rated by $3$ expert annotators. 
In cases of disagreement, two additional annotators are involved, and the final label is determined via discussion and majority voting. 
This protocol ensures consistent and high-quality annotations.


\subsection{Statistics}
Through rigorous video selection and question construction, we compile a dataset of $1,350$ videos, yielding $1,100$ annotated samples calibrated through multiple rounds of review. 
As shown in Figure~\ref{fig:stats}, the samples span five major categories, $20$ scenarios, and diverse durations, resolutions, and subject compositions. 
The AI-generated videos further cover a wide range of mainstream open-source and proprietary generation systems, allowing Artifact-Bench to capture diverse artifact distributions beyond a single model family or generation pipeline.

\section{Experiments}
\label{sec:Experiments}

\subsection{Evaluation Setup}

We evaluate a total of $19$ mainstream MLLMs, including $2$ cutting-edge proprietary models, $14$ open-source general-purpose models, and $3$ open-source specialized models designed for AI-generated videos detection.
Specifically, the proprietary models include Gemini 3.1 Pro~\cite{gemini-3.1-pro} and Gemini 3 Flash~\cite{gemini-3-flash}. 
The open-source general-purpose models include the Qwen3-VL series~\cite{qwen3-vl} (8B, 30B-A3B, and 32B), the InternVL3.5 series~\cite{internvl3.5} (8B, 30B-A3B, and 38B), Molmo2 8B~\cite{Molmo2}, MiMo-VL 7B~\cite{MiMo-VL}, and Keye-VL-1.5 8B~\cite{Keye-VL-1.5}.
The open-source specialized models include Skyra 7B~\cite{Skyra}, BusterX++~\cite{Busterx++}, and VideoVeritas 8B~\cite{VideoVeritas}.
To investigate whether reasoning-enhanced MLLMs improve the detection and assessment of artifacts in AI-generated videos, we further evaluate both instruction-tuned and reasoning-enhanced (i.e., thinking) variants of Qwen3-VL~\cite{qwen3-vl}, MiMo-VL~\cite{MiMo-VL}, and Skyra~\cite{Skyra}.
For all models, we adopt a default frame sampling rate of $5$ fps, with all other settings kept unchanged. Detailed experimental configurations are provided in Appendix~\ref{appendix:experiment_detail}.


\subsection{Main Results}

\begin{table*}[htbp]
\centering
\small
\setlength{\tabcolsep}{5pt}
\renewcommand{\arraystretch}{1.15}
\caption{\textbf{Evaluation results on \name.}
\textbf{RVAC} (Real vs. AI-generated Video Classification),
\textbf{PVRC} (Pairwise Video Realism Comparison),
and \textbf{AID} (Artifact Identification) denote the $3$ tasks in our benchmark.
Each task contains $3$ difficulty levels (\textbf{L1}–\textbf{L3}). 
\textbf{Avg} denotes the average accuracy across the $3$ difficulty levels.
\textbf{Total} denotes the overall score across all tasks.
The highest accuracy of each task (except Human Baseline) is highlighted in \colorbox{softgreen}{green}.
}
\label{tab:main_results}

\begingroup
\fontsize{6}{6}\selectfont  
\setlength{\tabcolsep}{4pt} 
\renewcommand{\arraystretch}{1.1}

\resizebox{\textwidth}{!}{

\begin{tabular}{lccccccccccccc}
\toprule
\multirow{2}{*}{\textbf{Model}} 
& \multicolumn{4}{c}{\textbf{Task 1: RVAC}} 
& \multicolumn{4}{c}{\textbf{Task 2: PVRC}} 
& \multicolumn{4}{c}{\textbf{Task 3: AID}} 
& \multirow{2}{*}{\textbf{Total}} \\

\cmidrule(lr){2-5}
\cmidrule(lr){6-9}
\cmidrule(lr){10-13}

& \textbf{L1} & \textbf{L2} & \textbf{L3} & \textbf{Avg}
& \textbf{L1} & \textbf{L2} & \textbf{L3} & \textbf{Avg}
& \textbf{L1} & \textbf{L2} & \textbf{L3} & \textbf{Avg}
& \\

\midrule

\rowcolor{softblue}
\multicolumn{14}{c}{\textbf{Proprietary Models}} \\

Gemini 3.1 Pro
& \colorbox{softgreen}{68.4} & \colorbox{softgreen}{76.5} & \colorbox{softgreen}{77.2} & \colorbox{softgreen}{74.0}
& 45.6 & 52.9 & 47.4 & 48.6
& \colorbox{softgreen}{19.3} & 6.4 & 3.8 & \colorbox{softgreen}{9.8}
& \colorbox{softgreen}{47.5} \\

Gemini 3 Flash
& 60.8 & 71.8 & 61.4 & 64.7
& 48.0 & \colorbox{softgreen}{57.5} & 47.4 & 50.9
& 8.6 & \colorbox{softgreen}{9.6} & \colorbox{softgreen}{11.3} & \colorbox{softgreen}{9.8}
& 43.8 \\

\midrule

\rowcolor{softpink}
\multicolumn{14}{c}{\textbf{Open-Source General-Purpose Models}} \\

Qwen3-VL 8B-Instruct
& 48.4 & 63.8 & 36.6 & 49.6
& 51.2 & 49.4 & 36.8 & 45.8
& 11.4 & 3.2 & 1.9 & 5.5
& 36.0 \\

Qwen3-VL 8B-Thinking
& 48.0 & 63.8 & 34.7 & 48.8
& 39.2 & 36.8 & 34.2 & 36.7
& 10.0 & 3.8 & 3.8 & 5.9
& 33.3 \\

Qwen3-VL 30B-A3B-Instruct
& 48.0 & 63.1 & 35.6 & 48.9
& 50.4 & 41.4 & 42.1 & 44.6
& 14.3 & 2.5 & 3.8 & 6.9
& 35.5 \\

Qwen3-VL 30B-A3B-Thinking
& 48.4 & 63.8 & 35.6 & 49.3
& 47.2 & 50.6 & 36.8 & 44.9
& 17.1 & 2.5 & 3.8 & 7.8
& 36.3 \\

Qwen3-VL 32B-Instruct
& 54.8 & 63.1 & 42.6 & 53.5
& 53.6 & 54.0 & 44.7 & 50.8
& 15.0 & 5.1 & 1.9 & 7.3
& 39.5 \\

Qwen3-VL 32B-Thinking
& 50.4 & 63.8 & 38.6 & 50.9
& 48.0 & 41.4 & 42.1 & 43.8
& 18.6 & 6.4 & 3.8 & 9.6
& 37.3 \\

InternVL3.5 8B
& 47.2 & 61.7 & 35.6 & 48.2
& 48.8 & 46.0 & 44.7 & 46.5
& 7.1 & 2.5 & 1.9 & 3.9
& 34.5 \\

InternVL3.5 30B-A3B
& 47.6 & 62.4 & 35.6 & 48.6
& 44.8 & 41.4 & 23.7 & 36.6
& 12.1 & 2.5 & 3.8 & 6.2
& 33.8 \\

InternVL3.5 38B
& 48.0 & 61.1 & 35.6 & 48.2
& 52.8 & 39.1 & 36.8 & 42.9
& 12.1 & 2.5 & 0.0 & 4.9
& 34.7 \\

Molmo2 8B
& 46.8 & 62.4 & 35.6 & 48.3
& 43.2 & 42.5 & 34.2 & 40.0
& 10.0 & 7.6 & 5.7 & 7.8
& 34.5 \\

MiMo-VL 7B-SFT
& 48.8 & 61.7 & 38.6 & 49.7
& 52.0 & 44.8 & \colorbox{softgreen}{52.6} & 49.8
& 5.7 & 2.5 & 0.0 & 2.8
& 35.4 \\

MiMo-VL 7B-RL
& 50.4 & 61.1 & 38.6 & 50.0
& 42.4 & 48.3 & 50.0 & 46.9
& 12.1 & 2.5 & 3.8 & 6.2
& 35.7 \\

Keye-VL-1.5 8B
& 48.8 & 61.7 & 35.6 & 48.7
& 48.8 & 37.9 & 47.4 & 44.7
& 5.0 & 1.3 & 1.9 & 2.7
& 33.8 \\

\rowcolor{softyellow}
\multicolumn{14}{c}{\textbf{Open-Source Specialized Models}} \\

Skyra 7B-SFT
& 47.2 & 63.8 & 36.6 & 49.2
& 19.2 & 23.0 & 21.1 & 21.1
& 10.0 & 3.2 & 3.8 & 5.7
& 29.4 \\

Skyra 7B-RL
& 51.2 & 62.4 & 40.6 & 51.4
& 31.2 & 27.6 & 18.4 & 25.7
& 8.6 & 3.2 & 5.7 & 5.8
& 32.0 \\

BusterX++ 7B
& 54.0 & 58.4 & 43.6 & 52.0
& 48.8 & 47.1 & 31.6 & 42.5
& 7.1 & 3.2 & 5.7 & 5.3
& 36.2 \\

VideoVeritas 8B
& 62.8 & 72.5 & 69.3 & 68.2
& \colorbox{softgreen}{60.8} & 56.3 & 42.1 & \colorbox{softgreen}{53.1}
& 16.4 & 3.2 & 3.8 & 7.8
& 46.0 \\

\midrule

\rowcolor{lightgray}
\multicolumn{14}{c}{\textbf{Human Baseline}} \\
Human Expert
& 95.6 & 92.6 & 90.1 & 93.6
& 88.0 & 86.2 & 81.6 & 86.4
& 82.9 & 79.0 & 77.4 & 80.3
& 87.7 \\

\bottomrule
\end{tabular}}

\endgroup

\end{table*}


We evaluate the performance of all models on \name~and display the accuracy in Table~\ref{tab:main_results}. To further analyze the preference alignment and performance gap between MLLMs and humans, we additionally invite four human experts to manually evaluate the benchmark.

\noindent\textbf{The experimental results reveal significant limitations of current MLLMs in artifact detection and identification scenarios.} Even Gemini~3.1~Pro achieves only an overall score of $47.5$ on \name, despite being the best-performing model.
It is worth noting that RVAC and PVRC are both binary decision tasks: RVAC requires a ``\texttt{Yes}'' or ``\texttt{No}'' answer, while PVRC requires selecting either ``\texttt{<Video A>}'' or ``\texttt{<Video B>}''. 
Thus, random guessing yields approximately $50\%$ accuracy. 
However, most MLLMs still fail to consistently surpass this baseline, especially at higher difficulty levels, indicating their limited ability to reliably recognize and compare realism-related artifacts in AI-generated videos.

\noindent\textbf{Existing MLLMs perform poorly on the AID task.} AID is substantially more challenging than RVAC and PVRC: instead of making a binary decision, models must select all valid artifact categories from six candidates, with multiple correct answers possible. Almost all models exhibit a dramatic performance drop on AID, with all models achieving less than $10\%$ average accuracy. These results suggest that although current MLLMs can partially recognize unrealistic videos at a coarse-grained level, they still struggle to explicitly analyze and explain the underlying causes of visual unrealism in AI-generated videos.

\noindent\textbf{A clear performance gap exists between proprietary and open-source models.} Overall, proprietary models consistently achieve stronger performance across all three tasks, indicating more robust capabilities in recognizing and reasoning about realism-related artifacts in AI-generated videos. However, despite their advantages, even the strongest proprietary models still exhibit a substantial gap compared with human experts. This result highlights the fundamental difficulty of artifact-aware video reasoning and suggests that current MLLMs remain far from reliably understanding the underlying causes of visual unrealism in AI-generated videos.


\subsection{Analysis and Findings}

\begin{figure*}
    \centering
    \includegraphics[width=0.98\linewidth]{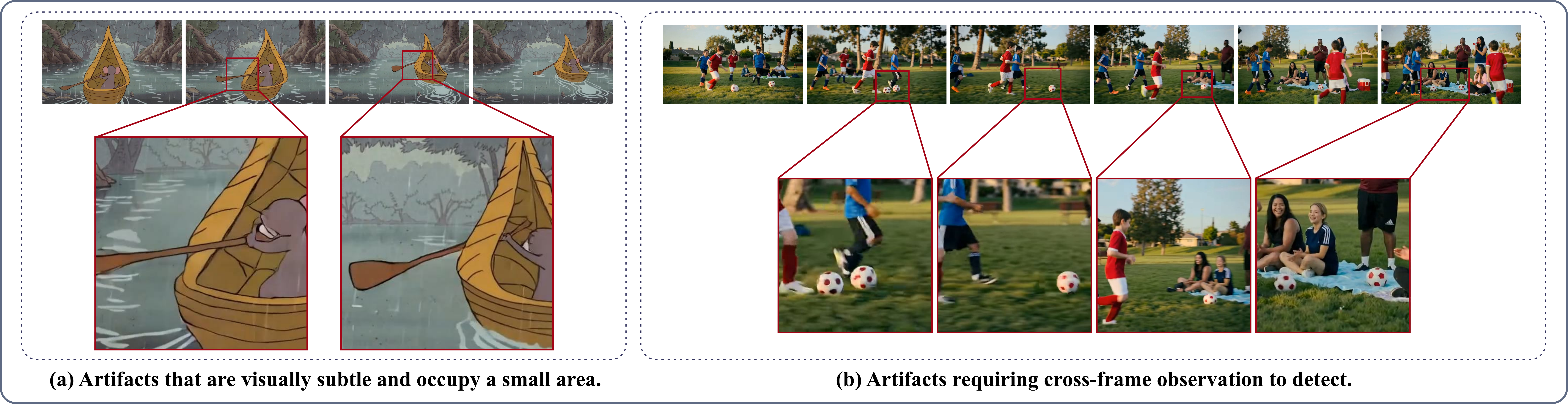}
    \caption{\textbf{Failure cases requiring fine-grained and temporal-spatial perception.} \textbf{(a)} The paddle penetrates the boat hull, while this artifact occupies only a small portion of the entire image; \textbf{(b)} the football changes from two balls to one and then back to two, requiring multi-frame object tracking.} 
    \label{fig:case_study}
\end{figure*}

\noindent\textbf{Fine-grained and temporal-spatial perception remain critical bottlenecks.} Figure \ref{fig:case_study} presents two representative failure cases for MLLMs. In Figure \ref{fig:case_study} (a), the artifact appears only in a small localized region, requiring fine-grained visual perception for accurate identification. In Figure \ref{fig:case_study} (b), the artifact is distributed across multiple frames, making temporal-spatial perception necessary for detection. These failure cases reveal fundamental limitations of current MLLMs in capturing subtle perceptual inconsistencies in AI-generated videos. For localized artifacts, the abnormal region often occupies only a very small portion of the frame and may be easily suppressed during visual token compression or global feature aggregation. As a result, models tend to focus on dominant semantic content while overlooking fine-grained structural abnormalities. Meanwhile, temporal-spatial artifacts are inherently more challenging because they cannot be identified from isolated frames alone. Detecting such inconsistencies requires models to jointly reason over object dynamics, motion continuity, and cross-frame structural consistency across long temporal contexts. However, current MLLMs often rely on sparse frame sampling and coarse temporal modeling, limiting their ability to capture subtle temporal evolution patterns. These observations suggest that reliable artifact-aware evaluation not only requires stronger semantic reasoning, but also demands substantially improved fine-grained perception and temporal-spatial modeling capabilities specifically tailored for generative artifact understanding.

\noindent\textbf{Scaling model size or enabling explicit reasoning does not necessarily improve artifact detection capability.}
For example, InternVL3.5-38B performs comparably to its 8B counterpart, while several thinking-enabled variants even underperform their instruction-tuned counterparts. This observation suggests that artifact detection and realism evaluation require capabilities beyond general semantic understanding and chain-of-thought reasoning. Unlike conventional multimodal reasoning tasks that primarily rely on high-level semantics or world knowledge, artifact-aware evaluation demands fine-grained perceptual sensitivity to subtle spatial-temporal inconsistencies, structural distortions, and abnormal motion patterns. Merely scaling model parameters or introducing generic reasoning processes may improve linguistic coherence and abstract reasoning ability, but does not necessarily enhance the model's ability to faithfully perceive artifacts

\noindent\textbf{Existing MLLMs exhibit a substantial mismatch between their artifact perception and human perceptual preferences.}
As the task difficulty progressively increases from L1 to L3, human performance consistently declines across all tasks, reflecting the increasing realism and perceptual ambiguity of the AI-generated videos. In contrast, the performance of current MLLMs often fluctuates irregularly or remains nearly unchanged across difficulty levels, rather than exhibiting a monotonic degradation trend aligned with human perception. In some cases, models even achieve comparable or higher accuracy on more challenging subsets. These observations suggest that current MLLMs do not reliably base their judgments on genuine artifact-aware perception. Instead, they may overly rely on superficial semantic cues, dataset biases, or shortcut correlations that are weakly related to perceptual realism itself.
This inconsistency reveals a fundamental limitation of current MLLMs in understanding video realism and generative artifacts. Although some models can partially distinguish AI-generated content under relatively easy settings, they fail to demonstrate stable human-aligned perceptual sensitivity as realism increases. Such misalignment substantially limits the reliability of MLLMs as general-purpose evaluators for AI-generated videos, particularly in applications requiring fine-grained realism assessment and artifact diagnosis.
More importantly, this issue may hinder the use of MLLMs as reward providers or automated judges for optimizing video generative models. Since reinforcement learning or preference optimization pipelines critically depend on stable and human-aligned reward signals, inaccurate artifact perception could encourage models to optimize toward superficial statistical patterns rather than genuinely improving perceptual realism. Our findings therefore highlight the urgent need for future MLLMs with stronger human-aligned artifact perception, temporal consistency understanding, and fine-grained realism reasoning capabilities.



\section{Conclusion}
\label{sec:Coclusion}
In this paper, we introduced \name, a benchmark for evaluating whether MLLMs can detect and diagnose artifacts in AI-generated videos. Through a three-level artifact taxonomy and three complementary tasks, Artifact-Bench provides a systematic evaluation from coarse-grained authenticity recognition to fine-grained artifact identification. Extensive experiments show that current MLLMs still struggle with artifact-level perception and reasoning. Moreover, model judgments are not always aligned with human preferences, limiting their reliability as evaluators or reward providers for video generative models. These findings highlight the need for future MLLMs with stronger fine-grained, temporal-spatial, and human-aligned realism understanding.

\clearpage
{
    \small
    \bibliographystyle{ieeenat_fullname}
    \bibliography{main}
}

\clearpage
\appendix
\section{Experiment Details}
\label{appendix:experiment_detail}

\subsection{Experimental Setup}
\label{app:experimental_setup}
For all evaluated models, we use a default video sampling rate of FPS$=5$ for video input. Due to context window limitations in some models and the high resolution of certain input videos, we additionally apply frame resizing when necessary to ensure feasible inference.

For decoding-related hyperparameters such as temperature, we prioritize the officially recommended settings for each model whenever available. For example, Gemini~3.1~Pro is evaluated using the official recommended configuration with $\texttt{temperature}=1.0$ and $\texttt{thinking\_level}=\texttt{"high"}$. Otherwise, models are evaluated using greedy decoding by default.

\subsection{Evaluation Prompt}

For reproducibility, we provide the prompt templates used for each task in \name~below.

\begin{tcolorbox}[title=Prompt Template for Task 1: Real vs. AI-Generated Video Classification (RVAC), colback=gray!5, colframe=gray!75]
\texttt{<Video>}
\\
\\
Determine whether the given video is generated by modern AI-based generative models (AIGC).
\\
\\
Videos created using traditional computer graphics (CG), animation pipelines, game engines, or professional rendering tools should NOT be classified as AIGC, even though they are not captured from the real world. This includes stylized or animated content, as well as game footage.
\\
\\
Such videos may exhibit visual artifacts due to artistic design, rendering limitations, compression, or real-time graphics constraints. However, these artifacts are fundamentally different from those introduced by AI generative models and should NOT be used as evidence for AIGC classification.
\\
\\
Your judgment should be based on the presence of artifacts that are characteristic of AIGC methods, arising from the limitations of generative models. 
\\
\\
Typical AIGC artifacts include, but are not limited to:
\\
- temporal inconsistency (e.g., flickering, unstable details across frames),
\\
- structural distortions (e.g., warped objects, inconsistent geometry),
\\
- unnatural motion or dynamics,
\\
- semantic incoherence (e.g., objects appearing/disappearing or morphing inconsistently),
\\
- abnormal visual appearance or texture anomalies (e.g., overly smooth, painterly, or “oily” rendering).
\\
\\
Do NOT rely solely on visual style (e.g., animation, rendering style, or game graphics) when making your judgment. Instead, focus on identifying artifact patterns that are indicative of AI-based generation.
\\
\\
If such AIGC-specific artifacts are clearly present, classify the video as "yes". Otherwise, classify it as "no".
\\
\\
Respond with "yes" if the video is AIGC, and "no" otherwise.
\end{tcolorbox}

\begin{tcolorbox}[title=Prompt Template for Task 2: Pairwise Video Realism Comparison (PVRC), colback=gray!5, colframe=gray!75]
Video A: \texttt{<Video A>}
\\
Video B: \texttt{<Video B>}
\\
\\
You are given two videos, <Video A> and <Video B>. Both videos are generated by modern AI-based generative models (AIGC).
\\
\\
Your task is to determine which video exhibits higher perceptual realism, i.e., which video contains fewer, less noticeable, or less severe artifacts introduced by generative models.
\\
\\
The comparison should be based on the presence and severity of AIGC-specific artifacts, rather than overall visual style or aesthetic preference.
\\
\\
Typical AIGC artifacts include, but are not limited to:
\\
- temporal inconsistency (e.g., flickering, unstable details across frames),
\\
- structural distortions (e.g., warped objects, inconsistent geometry),
\\
- unnatural motion or dynamics,
\\
- semantic incoherence (e.g., objects appearing/disappearing or morphing inconsistently),
\\
- abnormal visual appearance or texture anomalies (e.g., overly smooth, painterly, or “oily” rendering).
\\
\\
When making your decision:
\\
- Focus on the **severity, frequency, and perceptibility** of such artifacts.
\\
- A video with fewer and less perceptible artifacts should be considered more realistic.
\\
- Minor or barely noticeable artifacts are less important than severe or obvious ones.
\\
\\
Do NOT base your decision on:
\\
- stylistic differences,
\\
- resolution or sharpness alone,
\\
- color grading or lighting preferences,
\\
- high-level semantics or scene plausibility.
\\
\\
Do not be biased by video length, scene complexity, or content diversity.
\\
\\
Even if both videos contain artifacts, you must choose the one that is relatively more realistic.
\\
\\
Respond with:
\\
- "<Video A>" if <Video A> is more realistic,
\\
- "<Video B>" if <Video B> is more realistic.
\end{tcolorbox}

\begin{tcolorbox}[title=Prompt Template for Task 3: Artifact Identification (AID), colback=gray!5, colframe=gray!75]
\texttt{<Video>}
\\
\\
You are given a video that is generated by a modern AI-based generative model (AIGC).
\\
\\
For the following options, select all AIGC-specific artifacts that are clearly observable in the video.
\\
\\
Option:
\\
A. \texttt{[Option A]}
\\
B. \texttt{[Option B]}
\\
C. \texttt{[Option C]}
\\
D. \texttt{[Option D]}
\\
E. \texttt{[Option E]}
\\
F. \texttt{[Option F]}
\\
\\
Respond using the corresponding letter(s), separated by commas if multiple are selected (e.g., "A", "B,D", "A,C,E").
\end{tcolorbox}

\subsection{Answer Extraction Prompt}

We use the following prompt with Gemini 3.1 Pro to parse model responses and extract the final answers for accuracy evaluation.

\begin{tcolorbox}[title=Prompt Template for Task 1: Real vs. AI-Generated Video Classification (RVAC), colback=gray!5, colframe=gray!75]
You are an answer extraction system.
\\
\\
The original task is to determine whether a video is generated by an AI-based generative model (AIGC):
\\
- "yes" means the video is AI-generated.
\\
- "no" means the video is not AI-generated.
\\
\\
You are given a model response that may contain lengthy reasoning, analysis, self-corrections, or <think>...</think> blocks.
\\
\\
Your task is to extract only the model's final intended answer.
\\
\\
Model Response:
\\
\texttt{[Response]}
\\
\\
Rules:
\\
1. Ignore all reasoning, explanations, and intermediate analysis.
\\
2. Focus only on the final conclusion.
\\
3. The only valid outputs are:
\\
\hspace*{2em}- "yes"
\\
\hspace*{2em}- "no"
\\
4. Output exactly one valid answer without any additional text or explanation.
\\
5. If the response does not contain a clear final answer indicating whether the video is AI-generated, output: "Invalid"
\end{tcolorbox}

\begin{tcolorbox}[title=Prompt Template for Task 2: Pairwise Video Realism Comparison (PVRC), colback=gray!5, colframe=gray!75]
You are an answer extraction system.
\\
\\
The original task is to compare two AI-generated videos, <Video A> and <Video B>, and determine which video has higher perceptual realism, i.e., which video contains fewer or less severe AIGC-specific artifacts.
\\
\\
The only valid answers are:
\\
- "<Video A>"
\\
- "<Video B>"
\\
\\
You are given a model response that may contain lengthy reasoning, analysis, self-corrections, or <think>...</think> blocks.
\\
\\
Your task is to extract only the model's final intended answer.
\\
\\
Model Response:
\\
\texttt{[Response]}
\\
\\
Rules:
\\
1. Ignore all reasoning, explanations, and intermediate analysis.
\\
2. Focus only on the final conclusion.
\\
3. Output exactly one of the following:
\\
\hspace*{2em}- "<Video A>"
\\
\hspace*{2em}- "<Video B>"
\\
4. Do not output any additional text or explanation.
\\
5. If the response does not contain a clear final answer indicating which video is more realistic, output: "Invalid"
\end{tcolorbox}

\begin{tcolorbox}[title=Prompt Template for Task 3: Artifact Identification (AID), colback=gray!5, colframe=gray!75]
You are an answer extraction system.
\\
\\
The original task is to identify all AIGC-specific artifacts that are clearly observable in a given AI-generated video.
\\
\\
The candidate options are multiple-choice options labeled with letters (e.g., A-F). Multiple options may be correct.
\\
\\
You are given a model response that may contain lengthy reasoning, analysis, self-corrections, or <think>...</think> blocks.
\\
\\
Your task is to extract only the model's final intended answer.
\\
\\
Model Response:
\\
\texttt{[Response]}
\\
\\
Rules:
\\
1. Ignore all reasoning, explanations, and intermediate analysis.
\\
2. Focus only on the final conclusion.
\\
3. Extract only the selected option letters.
\\
4. If multiple options are selected, output them separated by commas (e.g., "A,C,E").
\\
5. Do not output any additional text or explanation.
\\
6. Only output valid option letters that appear in the final answer.
\\
7. If the response does not contain a clear final answer, output:
   "Invalid"
\end{tcolorbox}

\section{Benchmark Details}

\subsection{Representative Examples from \name}
\begin{figure*}[htbp]
    \centering
    \includegraphics[width=1\linewidth]{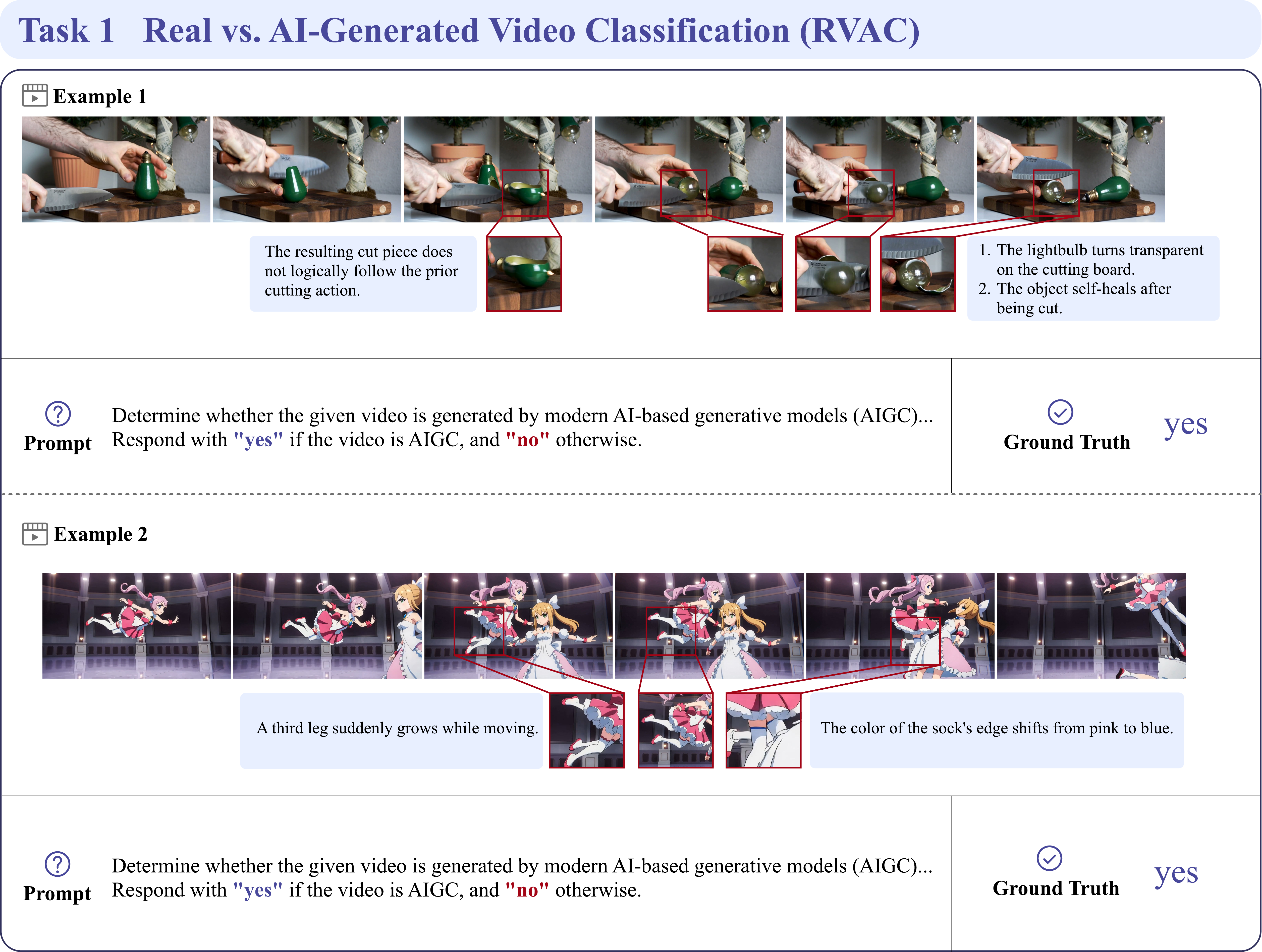}
    \caption{Representative examples for Task 1: Real vs. AI-Generated Video Classification (RVAC).}
    \label{fig:task1}
\end{figure*}

\begin{figure*}[htbp]
    \centering
    \includegraphics[width=1\linewidth]{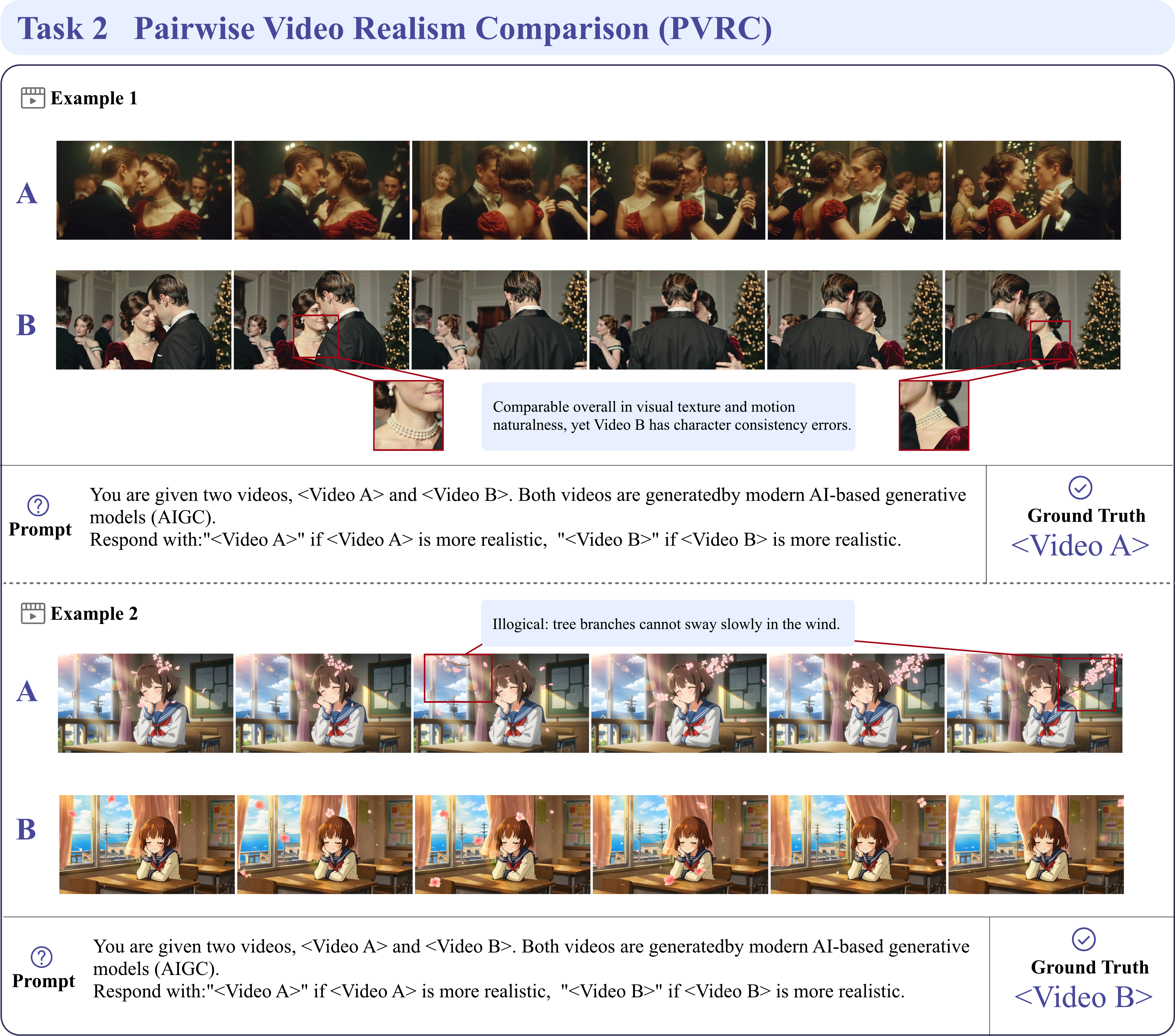}
    \caption{Representative examples for Task 2: Pairwise Video Realism Comparison (PVRC).}
    \label{fig:task2}
\end{figure*}

\begin{figure*}[htbp]
    \centering
    \includegraphics[width=1\linewidth]{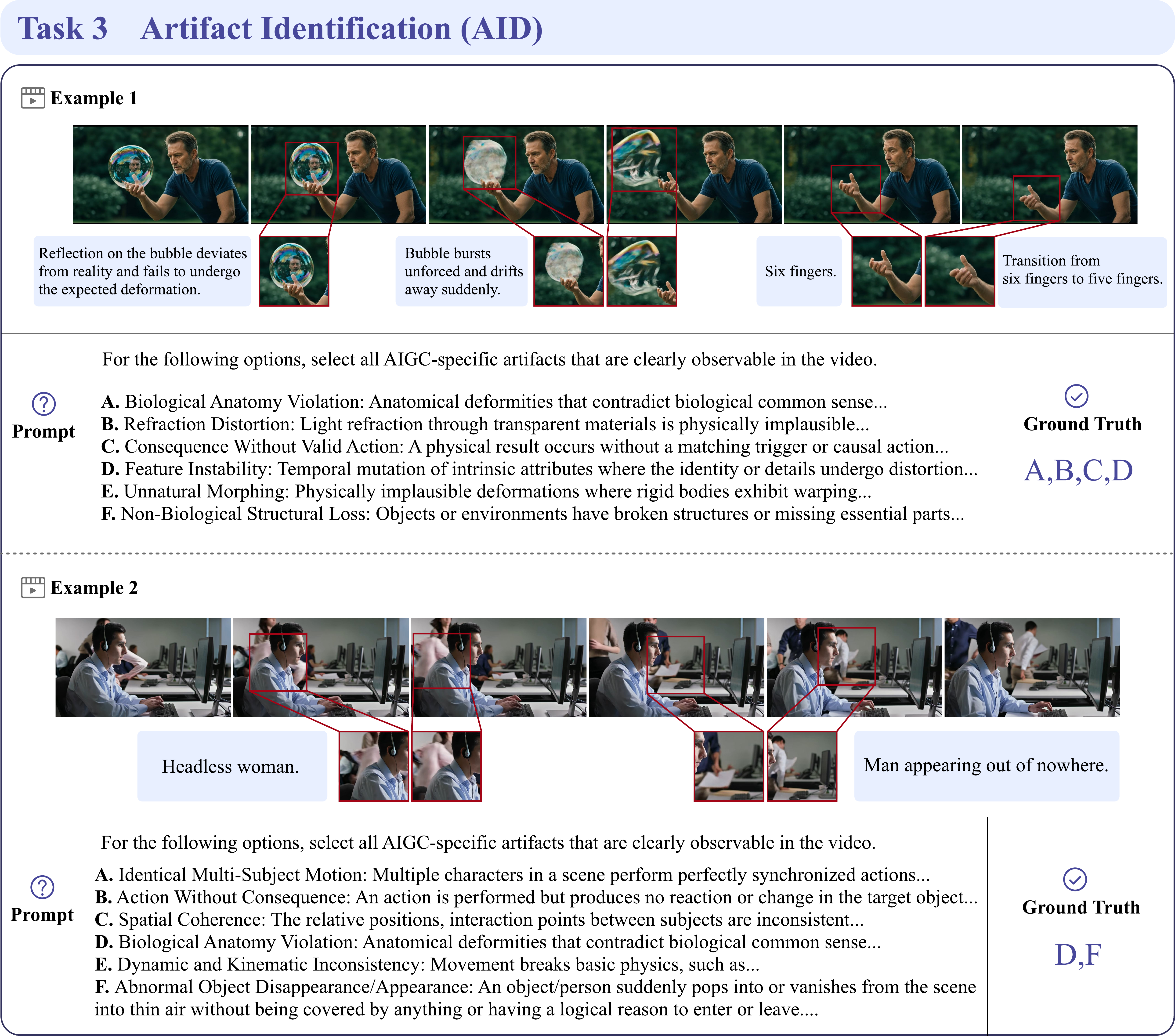}
    \caption{Representative examples for Task 3: Artifact Identification (AID).}
    \label{fig:task3}
\end{figure*}

In order to comprehensively convey the characteristics of
tasks in \name, two representative examples are presented for each task, as illustrated in Figures~\ref{fig:task1}, \ref{fig:task2}, and \ref{fig:task3}.

\subsection{Task Distribution}
Table~\ref{tab:task_distribution} shows the number of QA pairs of each difficulty level of each task in \name. 

\begin{table}[htbp]
\centering
\caption{Number of QA Pairs per task in RealVideo-Bench.}
\label{tab:task_distribution}
\renewcommand{\arraystretch}{1.15}
\begin{tabular}{p{0.55\linewidth} p{0.22\linewidth} c}
\toprule
\textbf{Task Category} & \textbf{Difficulty Level} & \textbf{\#QA} \\
\midrule

\multirow{3}{=}{Task 1: Real vs. AI-Generated Video Classification (RVAC)}
& L1 & 250 \\
& L2 & 149 \\
& L3 & 101 \\
& Total & 500 \\

\midrule

\multirow{3}{=}{Task 2: Pairwise Video Realism Comparison (PVRC)}
& L1 & 125 \\
& L2 & 87 \\
& L3 & 38 \\
& Total & 250 \\

\midrule

\multirow{3}{=}{Task 3: Artifact Identification (AID)}
& L1 & 140 \\
& L2 & 157 \\
& L3 & 53 \\
& Total & 350 \\

\midrule

Total & - & 1,100 \\

\bottomrule
\end{tabular}
\end{table}

\section{Limitations}
\label{sec:limitation}
Despite our efforts, Artifact-Bench still has limitations. Due to resource constraints, the number of human experts and the dataset scale can be further expanded. Future work will enlarge the benchmark with more diverse video sources, artifact types, and expert annotations, enabling more comprehensive and reliable evaluation of artifact-aware video understanding.

\section{Compute Resources}
\label{sec:ComputeResources}
All experiments were conducted on a distributed setup consisting of four identical machines, each equipped with 8 NVIDIA H800 GPUs and 1000 GiB of system memory. No additional compute beyond the reported experiments (excluding preliminary runs) is required to reproduce the main results.

\section{Impact Statement}
\label{sec:impact}
This paper presents work whose goal is to advance the field of Machine Learning. There are many potential societal consequences of our work, none of which we feel must be specifically highlighted here.


\end{document}